\documentclass{article}

\PassOptionsToPackage{numbers, sort&compress}{natbib}
\usepackage[final]{tackling_climate_workshop_style}

\usepackage[utf8]{inputenc} 
\usepackage[T1]{fontenc}    
\usepackage{hyperref}       
\usepackage{url}            
\usepackage{booktabs}       
\usepackage{amsfonts}       
\usepackage{nicefrac}       
\usepackage{microtype}      
\usepackage{xcolor}         
\usepackage{graphicx}
\usepackage{subcaption}
\usepackage{acronym}

\acrodef{DER}[DER]{{distributed energy resources}}
\acrodef{SciML}[SciML]{{Scientific machine learning}}
\acrodef{ML}[ML]{machine learning}
\acrodef{SMIB}[SMIB]{single-machine infinite-bus}
\acrodef{FNO}[FNO]{fourier neural operator}

\usepackage{enumitem,amssymb, amsmath}
\newlist{todolist}{itemize}{2}
\setlist[todolist]{label=$\square$}
\usepackage{pifont}

\usepackage{todonotes}

\title{Operator Learning for Power Systems Simulation}

\author{
Matthew Schlegel \\
Schulich School of Engineering \\
University of Calgary \\
\texttt{matthew.schlegel@ucalgary.ca} \\
\And
Matthew E. Taylor \\
Deparment of Computing Science \\
University of Alberta \\
\texttt{mtaylor3@ualberta.ca} \\
\And
Mostafa Farrokhabadi \\
Schulich School of Engineering \\
University of Calgary \\
\texttt{mostafa.farrokhabadi@ucalgary.ca} \\
}

\begin{document}

\maketitle

\begin{abstract}
Time domain simulation, i.e., modeling the system's evolution over time, is a crucial tool for studying and enhancing power system stability and dynamic performance. However, these simulations become computationally intractable for renewable-penetrated grids, due to the small simulation time step required to capture renewable energy resources' ultra-fast dynamic phenomena in the range of 1-50 microseconds. This creates a critical need for solutions that are both fast and scalable, posing a major barrier for the stable integration of renewable energy resources and thus climate change mitigation. This paper explores operator learning, a family of machine learning methods that learn mappings between functions, as a surrogate model for these costly simulations. The paper investigates, for the first time, the fundamental concept of simulation time step-invariance, which enables models trained on coarse time steps to generalize to fine-resolution dynamics. Three operator learning methods are benchmarked on a simple test system that, while not incorporating practical complexities of renewable-penetrated grids, serves as a first proof-of-concept to demonstrate the viability of time step-invariance. Models are evaluated on (i) zero-shot super-resolution, where training is performed on a coarse simulation time step and inference is performed at super-resolution, and (ii) generalization between stable and unstable dynamic regimes. This work addresses a key challenge in the integration of renewable energy for the mitigation of climate change by benchmarking operator learning methods to model physical systems.
\end{abstract}

\section{Introduction}

The increasing penetration of renewable and distributed energy resources has introduced novel dynamic phenomena to the power grid, creating new challenges for system stability analysis. Time domain simulation has been used as a reliable tool for studying system stability. In general, power systems time domain simulation involves the numerical integration of nonlinear differential-algebraic equations \cite{kundur2007power}, which are computationally intensive. This bottleneck has rendered system-wide simulation computationally intractable as renewable generation grows, given the need for fine-grained integration time steps to capture novel dynamics associated with inverters \cite{pfenninger2014energy, lopion2018review}. Recent grid incidents, such as the April 2025 massive blackout in Spain and Portugal, underscore these limitations and the need for scalable, real-time analysis tools. Overcoming this computational barrier is crucial to the stable integration of renewable energy sources as mandated by climate change mitigation policies.

Prior works on \ac{ML}-based modeling for power system simulation can be broadly categorized into two approaches. First are methods that use classifiers to determine system stability status and/or stability margin post-disturbance \cite{ wehenkel1989artificial, he2013robust, yu2018intelligent, gupta2019online, zhu2020hierarchical}. These are not in the scope of this paper due to their inability to be used as surrogate models for time domain simulation. While helpful for stability prediction, they cannot be used for control design and stability enhancement. Second are approaches that have used surrogate models to produce trajectories from initial conditions \cite{li2020machinelearningbased, moya2023deeponetgriduq, xiao2023feasibility, bossart2025acceleration, cheng2025machinelearningreinforced}. Surrogate modeling can be further subdivided into two approaches. The first replaces a component in the full simulation with a model that approximates its differential equations \cite{xiao2023feasibility, bossart2025acceleration}. 
This has the advantage of being incorporated directly into existing simulation frameworks, but necessitates significant complexity to create an interface to integrate the machine learning-based component into the differential equations for the rest of the system \cite{bossart2025acceleration}. Another approach, which is the focus of this paper, is to devise a surrogate data-driven model for the entire system. However, a surrogate model of the full system requires extensive training, verification, and generalization tests.

In this paper, a surrogate model is used to learn a function, in this case an operator, that maps from an initial disturbed trajectory of a test system. 
Operator learning has emerged in scientific \ac{ML} offering discretization invariance, and thus the potential to generalize across resolutions. 
The immediate contributions of this study are: (i) a comparison of three operator learning methods on zero-shot super-resolution generalization, and (ii) an analysis of the generalization between stable and unstable regimes. The long-term contributions of the paper are: (i) providing a path for stable integration of renewable energies through a scalable and computationally tractable analysis tool, and (ii) tackling machine learning for modeling complex physical systems. 

\section{Operator Learning}

Operator learning approximates a mapping between two infinite-dimensional spaces\footnote{The interested reader should refer to \cite{lu2021learning} and \cite{kovachki2023neural} for more details on the operator learning problem setting.} \cite{lu2021learning, kovachki2023neural, azizzadenesheli2024neural}. Methods in operator learning enable discretization-invariant predictions through algorithmic and architectural designs.
This property allows models trained on coarse time domain simulations ($\Delta t = 100\,\text{ms}$) to potentially generalize to finer simulation time steps ($\Delta t = 50\,\mu\text{s}$), reducing the need for computationally-intensive high-resolution simulation to generate training data for a surrogate model. In this paper, we learn an operator $G$ that maps from an initial trajectory $t\in[0, \tau]$ to a future expected trajectory generated from an ordinary differential equation $t\in[\tau_i, T]$.

\textbf{Deep Operator Networks (DeepONets)} \cite{lu2021learning} use two neural network branches: a branch network processes the input trajectory, and a trunk network encodes the output query time. Their outputs are combined via an inner product to produce the predicted trajectory value.

\textbf{Fourier Neural Operators (FNOs)} \cite{kovachki2023neural, li2020fourier, kossaifi2024multigrid} project inputs into a Fourier basis, apply learned spectral multipliers, and transform back to the time domain. Tensorization reduces parameter count while preserving accuracy compared to standard FNOs. This spectral representation can capture smooth temporal structures and support resolution changes.

\textbf{Latent Neural ODEs (LNODEs)} \cite{chen2018neural} encode the input trajectory into a latent state, evolve it through a neural ODE, and decode it back to the original space. We use an autoencoder variation rather than a variational autoencoder for a fairer comparison. The fixed-step (Adams) and adaptive-step (Dormand–Prince–Shampine) solvers are evaluated.

The rationale for selecting the above methods is as follows: neural ODEs have been used in the prior art to model independent components of a power system to integrate into a traditional simulation \cite{xiao2023feasibility,bossart2025acceleration}. DeepONets have been applied to post-disturbance power system modeling \cite{moya2023deeponetgriduq}. FNOs, while suitable for the task, have not been used for power system modeling. This paper presents the first step towards the application of FNOs to power system modeling.

\section{Experimental Setup}

\subsection{SMIB Test System}
\label{sec:smib_physics}

This study focuses on a \ac{SMIB} test system, which is a single synchronous machine connected to a stiff, i.e., fixed voltage and frequency, source via a transmission line \citep{saadat1999power}. Without the loss of generality, the generator angle dynamics are used for this study, associated with a classic stability phenomenon: if a generator's mismatch between input mechanical and output electrical powers exceeds a critical threshold after a disturbance, i.e., any change that alters the steady-state or normal dynamic behavior of the system, the machine can no longer maintain a stable exchange of power with the grid. This instability is a well-known real-world phenomenon in power systems and continues to pose practical challenges. As such, the \ac{SMIB}, despite its simplicity, serves as a suitable candidate to explore new proof-of-concept simulation approaches; it is simple enough to isolate the fundamental dynamics, yet rich enough to reflect practical stability issues.

A synchronous machine angle dynamics are governed by the following equation:
\begin{equation}
    \frac{\partial^2 \delta}{\partial t^2} = \frac{\pi f_0}{H} \left(P_m - D \frac{\partial \delta}{\partial t} - \frac{|E| |V|}{X} \sin\delta\right),
\end{equation}
where $\delta(t)$ is the rotor (power) angle in electrical radians, $P_m$ is the mechanical power input per-unitized on the machine base, $D$ is the damping coefficient in per unit power per radians/s, $E$ and $V$ are the machine and stiff source voltages per unitized on a common base, $X$ is the reactance of the transmission line between the two sources in per unit, $H$ is the inertia constant in seconds, and $f_0$ is the nominal frequency in Hz. At steady state, the pre-disturbance rotor angle is $\delta_0 = \sin^{-1}(P_m / P_{m}^{\max})$, where $P_{m}^{\max} = |E||V| / X$ is the maximum transferable electrical power. A disturbance at $t=0$ changes the mechanical input from $P_m$ to $P_{m1}$. In the undamped case ($D=0$), the maximum post-disturbance mechanical input preserving stability is:
\begin{align}
P_{m1}^{\max} &= P_m^{\max} \sin(\pi - \delta_{\max}) \\
0 &= (\delta_{\max} - \delta_0)\sin(\delta_{\max}) + \cos(\delta_{max}) - \cos(\delta_0).
\end{align}
where $\delta_{\max}$ is determined by Equation (3). Integration is performed using Runge-Kutta of order 5(4).

\subsection{Data Generation and Model Training}

\ac{SMIB} trajectories are generated in stable and unstable regimes by sampling the $P_m \sim \mathcal{U}[0, 2]$, and damping $D\sim[0.0, 0.135]$. For stable regimes, we sample $P_{m1} \sim \mathcal{U}[0.0, P_{m1}^{\max}]$ assuming $D=0$ to avoid mislabeled trajectories. 
For unstable trajectories, the lower bound of the unstable range is approximated using a 100-iteration bisection search over the range $[P_{m1}^{\max},P_{\max}]$ \cite{epperson2013introduction}. \ac{SMIB} parameters are resampled if $P_{\max}$ does not result in an unstable trajectory.
Other parameters are taken from an example in \cite{saadat1999power}: $E=1.35$, $V=1.0$, $X=0.65$, $H=9.94$, and  $f_0 = 60$. These may not reflect a practical power system but their values are unlikely to change the conclusions. Each dataset had 8000 training trajectories and 1000 validation trajectories. The input sequence covers $t\in[0, 0.2]\,\text{s}$  and the target covers $t\in [0.3, 3.1]\,\text{s}$. The input features $\delta$ is clipped to a max value of $\pi$ and $\mathbf{x}=[\delta, \omega]$ are normalized such that $\delta \in [0,1]$ and $\omega \in [-1, 1]$ using ranges in the training set.

Each model is tuned through 120 trials using a Bayesian hyperparameter optimization method \cite{bergstra2011algorithms, akiba2019optuna} with a small number of epochs (20 for the LNODE and FNO and 200 for the DeepONets). All the models have approximately $\sim 700,000$ parameters to ensure a fair comparison. Using the best hyperparameter settings, each model was trained to obtain a reasonable validation error, or until the validation error stopped decreasing. 
The LNODEs and FNOs took 60 epochs, while the DeepONets took 600 epochs. 
All methods used the H1 Sobolev norm loss for training (see \cite{kovachki2023neural} for details).

\section{Results}

\begin{figure}[t!]
    \centering
    \includegraphics[width=\linewidth]{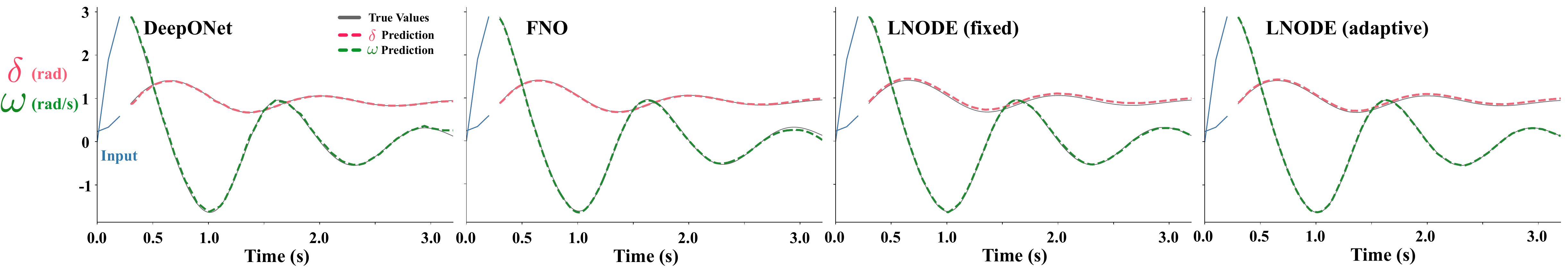}
    \caption{\textbf{Zero-Shot Super-Resolution}:
    This figure illustrates the true time series of the system angle ($\delta$) and frequency ($\omega$) dynamics in solid lines and the operators' predicted trajectories in dotted lines. The plots start with the segment of the true trajectory in a solid blue line used as the input.}
    \label{fig:smib_trajectories}    
\end{figure}

\textbf{Zero-Shot Super-Resolution:} Models are trained on data generated on a coarse time step ($100\,\text{ms}$) and compared on their generalization to data generated on a fine time step ($50\,\mu s$) without retraining. Results are reported in Table \ref{tab:super_resolution} with example trajectories in Figure \ref{fig:smib_trajectories}. Overall, the LNODE outperformed both DeepONets and FNOs, with the LNODE achieving the lowest change in performance between the resolutions. While the FNOs outperformed the other approaches in absolute RMSE on the original training resolution, they had the highest drop in performance for the higher resolution.

\begin{table}[h!]
\centering
\begin{tabular}{l|c|c|c}
\textbf{Model Name} & {\textbf{ $\Delta t=100ms$ (RMSE) }} & {\textbf{ $\Delta t=50\mu s$ (RMSE)}} & {\textbf{Percent Difference}}  \\
\hline
DeepONet & $0.0220\pm0.0001$  & $0.0348\pm0.0002$ & $45.2 (39.0, 51.31) $  \\
FNO &     $0.0186\pm0.0001$  & $0.0302\pm0.0001$ & $47.5 (39.9, 54.8)$ \\
LNODE (Fixed) &$0.0280\pm0.0006$  & $0.0305\pm0.0006$ & $8.6 (0.5, 33.6)$ \\
LNODE (Adaptive) &$0.0275\pm0.0003$  & $0.0296\pm0.0003$ & $7.3 (0.4, 19.7)$
\end{tabular}\vspace{0.1 cm}
\caption{\textbf{Zero-Shot Super-Resolution}: Values are reported over 200 test trajectories and 20 independent runs. Both resolution datasets use the same SMIB parameters. Bootstrap confidence intervals are used for the percent difference and standard errors are reported for RMSE values.}
\label{tab:super_resolution}
\vspace{-0.4cm}
\end{table}

\begin{figure}[t!]
    \centering
    \includegraphics[width=\linewidth]{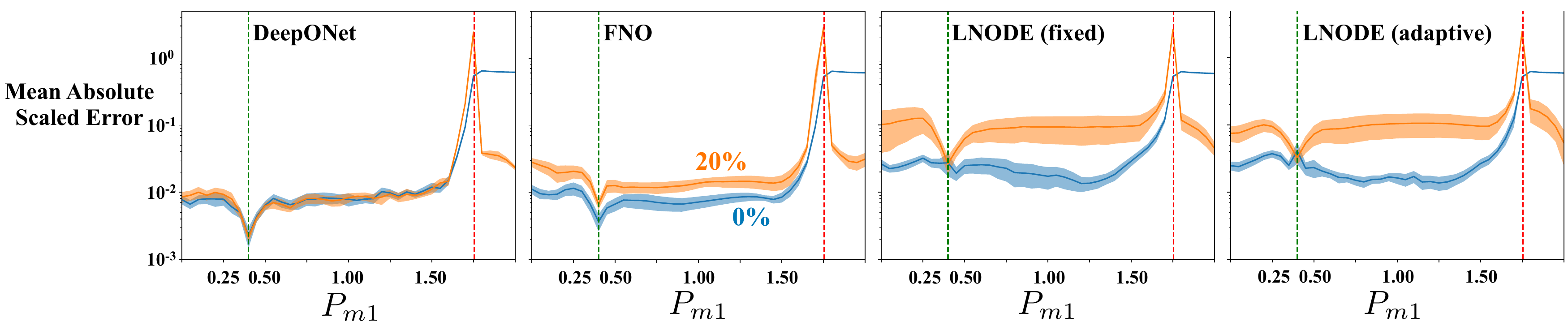}
    \caption{\textbf{Generalization across different dynamical regimes}: Training data has $0\%$ (blue) and $20\%$ (orange) unstable trajectories. The green dotted line is where $P_{m1} = P_m$, i.e., no disturbance, and the red dotted line is the critical point where trajectories become unstable. The mean absolute scaled error is calculated using a baseline-prediction of the previous time step over 20 runs.}
    \label{fig:smib_gen}
    \vspace{-0.4cm}
\end{figure}

\textbf{Generalization across different dynamic regimes: } The models are trained on two datasets: one with $0\%$ unstable trajectories and one containing $20\%$ unstable trajectories for training. The models are compared on their performance for a single \ac{SMIB} system, $P_m=0.4$ and $D=0.05$, and a sweep over the post-disturbance mechanical power $P_{m1}$. 
The system loses stability at a critical threshold marked by the red dashed line in Figure \ref{fig:smib_gen}. As seen in Figure \ref{fig:smib_gen}, the methods struggle to generalize to dynamical regimes not covered by the training data; this is consistent with prior observations \cite{subramanian2023foundation}. When trained with a mixture of stable and unstable trajectories, all the methods are able to capture the unstable region's dynamics, while the DeepONet is the only method to perform similarly in the stable regime when the training dataset includes both stable and unstable trajectories.

\section{Discussion}

The main limitation of this work is the simplicity of the \ac{SMIB}, which does not capture the complex dynamics of renewable-penetrated grids. Although all methods performed well, comprehensive modeling and testing on practical test systems are needed to draw generalizable conclusions. For example, on this simple test system, the FNO was shown to outperform alternatives in absolute error in the two evaluation tasks, but the DeepONets and LNODE approaches are likely to be competitive with different inductive biases \cite{lu2022comprehensive}; more tests are needed to prove this. This study provides a controlled setting to establish the feasibility of time step-invariant operator learning. This capability addresses a major bottleneck in renewable-penetrated power system analysis through time domain simulation, ultimately supporting the transition to low-carbon grids needed for climate change mitigation.

\textbf{GitHub Repository:} \url{https://github.com/mkschleg/NOForPSSim.git}

\begin{ack}
This research was done in the Data+Grid Research and Innovation (DGRI) Lab at the University of Calgary and the Intelligent Robot Learning (IRL) Lab at the University of Alberta. This research was funded by the Natural Sciences
and Engineering Research Council (NSERC) of Canada Grant RGPIN-2024-03955 and supported in part by Alberta Innovates; Alberta Machine Intelligence Institute (Amii); a Canada CIFAR AI Chair, Amii; Digital Research Alliance of Canada; and Mitacs.
\end{ack}

\bibliography{main}
\bibliographystyle{abbrv}

\end{document}